\documentclass{article}     % uncomment this line when writing your paper

\usepackage{arxiv}
\usepackage[utf8]{inputenc} % allow utf-8 input
\usepackage[T1]{fontenc}    % use 8-bit T1 fonts
\usepackage{hyperref}       % hyperlinks
\usepackage{url}            % simple URL typesetting
\usepackage{booktabs}       % professional-quality tables
\usepackage{amsfonts}       % blackboard math symbols
\usepackage{nicefrac}       % compact symbols for 1/2, etc.
\usepackage{microtype}      % microtypography
\usepackage{lipsum}
\usepackage{graphicx}
\usepackage{caption}
\usepackage{subcaption}
\usepackage{multirow}
\title{Describing Robots from Design to Learning: \\Towards an Interactive Lifecycle Representation of Robots\thanks{A preprint submitted to the \textbf{IEEE ICRA 2024}.}}
\author{
    {Nuofan Qiu}\\
    School of Design\\
    Southern University of Science and Technology\\
    Shenzhen, China 518055\\
    \And
    {Fang Wan*}\\
    School of Design\\
    Southern University of Science and Technology\\
    Shenzhen, China 518055\\
    \texttt{wanf@sustech.edu.cn} \\
    \And
    {Chaoyang Song*}\\
    Department of Mechanical and Energy Engineering\\
    Southern University of Science and Technology\\
    Shenzhen, China 518055\\
    \texttt{songcy@ieee.org} \\
}
\begin{document}
\maketitle
%%%%%%%%%%%%%%%%%%%%%%%%%%%%%%%%%%%%%%%%%
\begin{abstract}

    The robot development process is divided into several stages, which create barriers to the exchange of information between these different stages. We advocate for an interactive lifecycle representation, extending from robot morphology design to learning, and introduce the role of robot description formats in facilitating information transfer throughout this pipeline. We analyzed the relationship between design and simulation, enabling us to employ robot process automation methods for transferring information from the design phase to the learning phase in simulation. As part of this effort, we have developed an open-source plugin called ACDC4Robot for Fusion 360, which automates this process and transforms Fusion 360 into a user-friendly graphical interface for creating and editing robot description formats. Additionally, we offer an out-of-the-box robot model library to streamline and reduce repetitive tasks. All codes are hosted open-source. (\url{https://github.com/bionicdl-sustech/ACDC4Robot})
    
\end{abstract}
%%%%%%%%%%%%%%%%%%%%%%%%%%%%%%%%%%%%%%%%%
%%%%%%%%%%%%%%%%%%%%%%%%%%%%%%%%%%%%%%%%%
% \newpage
% \tableofcontents        % You can remove TOC before final submission, but keep it while preparing the manuscript to remind yourself on the paper structure
% \newpage
%%%%%%%%%%%%%%%%%%%%%%%%%%%%%%%%%
\section{INTRODUCTION}
\label{sec:Intro}
%%%%%%%%%%%%%%%%%%%%%%%%%%%%%%%%%

    As autonomous machines capable of interacting with the real world, various types of robots, such as wheeled mobile robots, quadrupedal robots, and humanoid robots, are emerging in domestic, factory, and other environments to collaborate with humans or accomplish tasks independently. The morphology of a robot is the essential factor that most directly affects the robot's configuration space, thereby determining the robot's function \cite{feldotto2022neurorobotics}. Robot morphology is primarily determined during the design process, thanks to the development of computer-aided design (CAD) technology, which makes it cost-effective, time-saving, and efficient compared to the manufacturing process.

    Beyond robot morphology, learning has become an essential topic in robotics because it enables robots to achieve complex tasks and, thus, better interact with the environment. However, training robots in hardware may lead to failures or damage, making it expensive and time-consuming. Simulation provides a more cost-effective and safer way to develop robots. Moreover, robot simulators incorporate domain randomization techniques that increase the exploration of the state-action space, facilitating the transfer of knowledge learned in simulation to real robots \cite{muratore2022robot}. All robot simulators construct simulation instances from robot models derived from \textit{robot description formats}. 

    \begin{figure}[t]
        \centering
        \includegraphics[width=0.7\columnwidth]{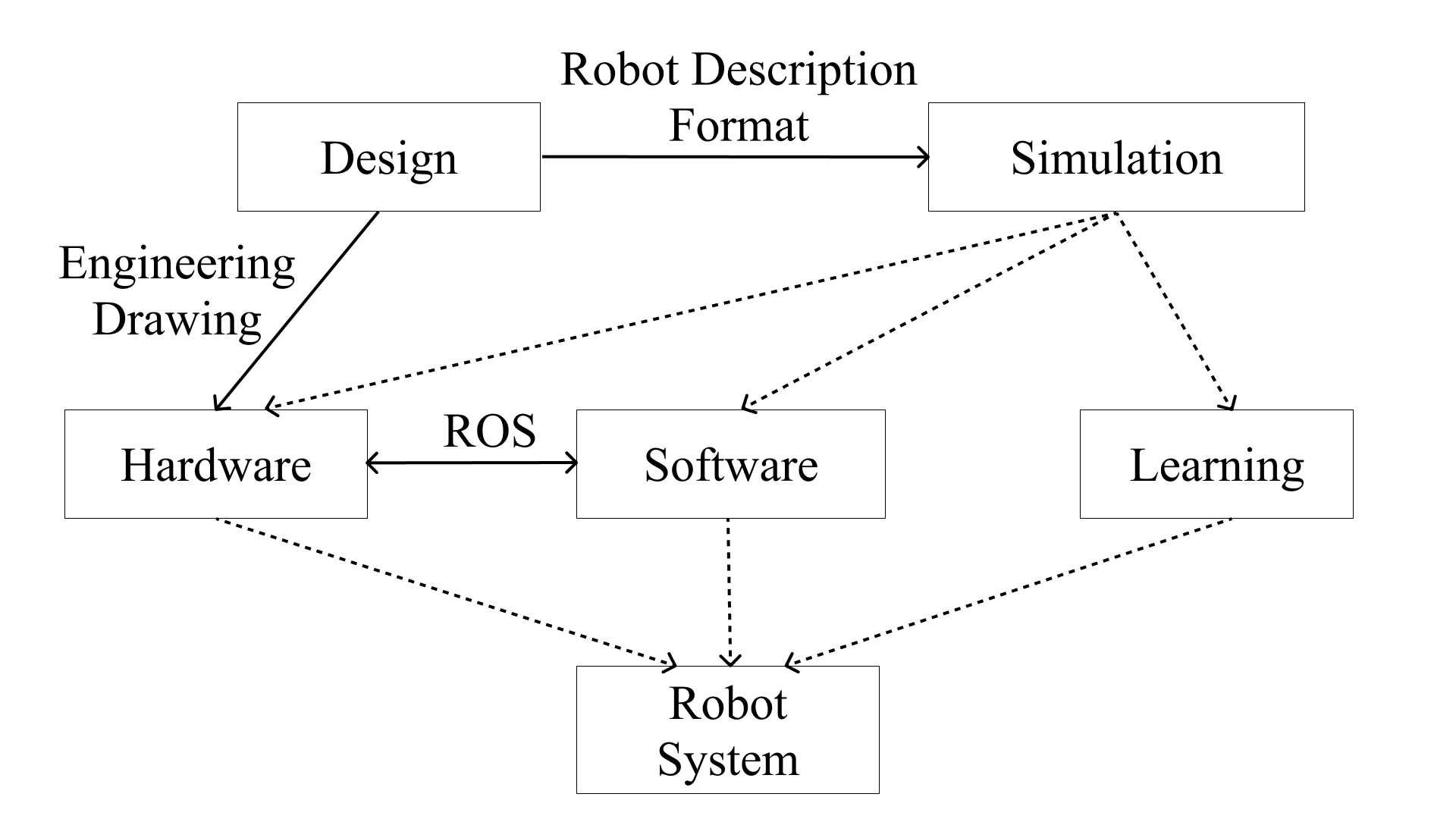}
        \caption{
        \textbf{Robot System Development Framework.}
        Robot system development is divided into several phases. The exchange of information between these different phases consumes a significant amount of developer time and energy. Many tools have been developed for this purpose; for example, design information is transferred to hardware through engineering drawings during manufacturing, and ROS is developed to facilitate information transfer between software and hardware.
        }
        \label{fig:Robot-Framework}
    \end{figure}

    Robot Description Format (RDF) is a class of formats that can describe the robot system in a structured manner following a set of rules. RDF contains information about the robot system, including kinematics, dynamics, actuators, sensors, and the environment with which the robot can interact. RDF can transfer information about the robot from the design phase to simulation; thus, it can be seen as the interface between robot morphology design and robot learning in a simulated environment.

\subsection{File Formats from Design to Learning}

    Several file formats are used in robot morphology design and learning in a simulation environment. These file formats have specific features tailored to different application scenarios, hindering process interoperability. Various file formats make it challenging to transfer information from the design phase to the learning process in simulation.

    In contemporary practice, robot morphology is typically designed using CAD software. File formats in the CAD field can be categorized into neutral and native formats. Neutral file formats adhere to cross-platform compatibility standards, including STEP files (.stp, .step), IGES files (.igs, .ige), COLLADA, and STL. Native file formats are platform-specific and contain precise information optimized for the respective platform, examples of which include SolidWorks (.sldprt, .sldasm), Fusion 360 (.f3d), Blender (.blend), and many others.

    Several robot description formats are used in robot simulation. The most common format is the Unified Robotics Description Format (URDF), which is supported by various robot simulators, including PyBullet, Gazebo, and MuJoCo. SDFormat is natively supported by Gazebo and partially supported by PyBullet. MuJoCo natively supports MJCF and is also supported by Isaac Sim and PyBullet. Other robot description formats resemble native formats specific to particular simulators than URDF. For example, the CoppeliaSim file is designed for use with CoppeliaSim, and WBT is used in Webots.

\subsection{A Brief Historical Review of Robot Description Formats}

    Robot Description Formats provide information for modeling the robot system and are widely used in robot simulators. Currently, research resources on robot description formats are limited, with most of the relevant information available only on their respective websites and forums, making research challenging. The authors in \cite{ivanou2021robot} compared existing formats and summarized their main advantages and limitations. Here, we offer a concise historical perspective on robot description formats to enhance understanding.

\subsubsection{Before Unified Robot Description Format (URDF)}

    Research on robot modeling predates the concept of a robot description format by a considerable margin. Denavit and Hartenberg formulated a convention using four parameters to model robot manipulators in 1955 \cite{denavit1955kinematic}, which is still widely used in robotics. With the advent of computer simulation, robots can be defined using programming languages with variables \cite{vzlajpah2008simulation}. While it is theoretically possible to describe a robotic system through a programming language's variables and data structures, the reliance on programming language features can make it cumbersome to exchange robot system information across different platforms for various purposes. Therefore, representing robot system information in a unified, programming language-independent manner will facilitate interchangeability across other platforms and enhance development efficiency. Park et al. \cite{hwan2007xml} discussed XML-based formats, which can describe robots due to XML's convenience in delivering information.

\subsubsection{URDF, SDFormat, and Others}

    While developing a personal robotics platform, the idea of creating a ``Linux for robotics'' came to the minds of Eric Berger and Keenan Wyrobek \cite{wyrobek2008towards}. With the first distribution of ROS—ROS Mango Tango—released in 2009, URDF was simultaneously introduced. URDF is an XML-based file format that enhances readability and describes robot links' information, including kinematics, dynamics, geometries, and robot joints' information organized in a tree structure. URDF universally models robots, making them suitable for visualization, simulation, and planning within the ROS framework.

    With the growing popularity of ROS, URDF has become a widely used robot description format supported by various simulation platforms, such as PyBullet, MuJoCo, and Isaac Sim, among others. However, an increasing number of roboticists have recognized the limitations and issues of URDF, such as its inability to support closed-loop chains. The community has endorsed proposals like URDF2 to address these concerns\footnote{https://sachinchitta.github.io/urdf2/}. The problems stemming from URDF's design may become increasingly challenging to resolve over time due to the diminishing activity in its development (the repository's\footnote{https://github.com/ros/urdf} update frequency has become very low). Therefore, new formats can draw upon URDF's experience to avoid such issues from the outset and expand their ability to describe a broader range of scenarios.

    Rosen Diankov et al. \cite{diankov2011collada} promoted an XML-based open standard called COLLADA, which allows for complex kinematics with closed-loop chains. SDFormat (Simulation Description Format) was initially developed as part of the Gazebo simulator and separated from Gazebo as an independent project to enhance versatility across different platforms. SDFormat is also an XML-based format that shares a similar grammar with URDF but extends its ability to describe the environment with which the robot interacts. Furthermore, SDFormat is actively developing, making it more responsive to future robotics needs. MJCF is another XML-based file format initially used in the MuJoCo simulator. It can describe robot structures, including kinematics, dynamics, and other elements like sensors and motors.

\begin{figure}[htbp]
        \centering
        \includegraphics[width=0.7\columnwidth]{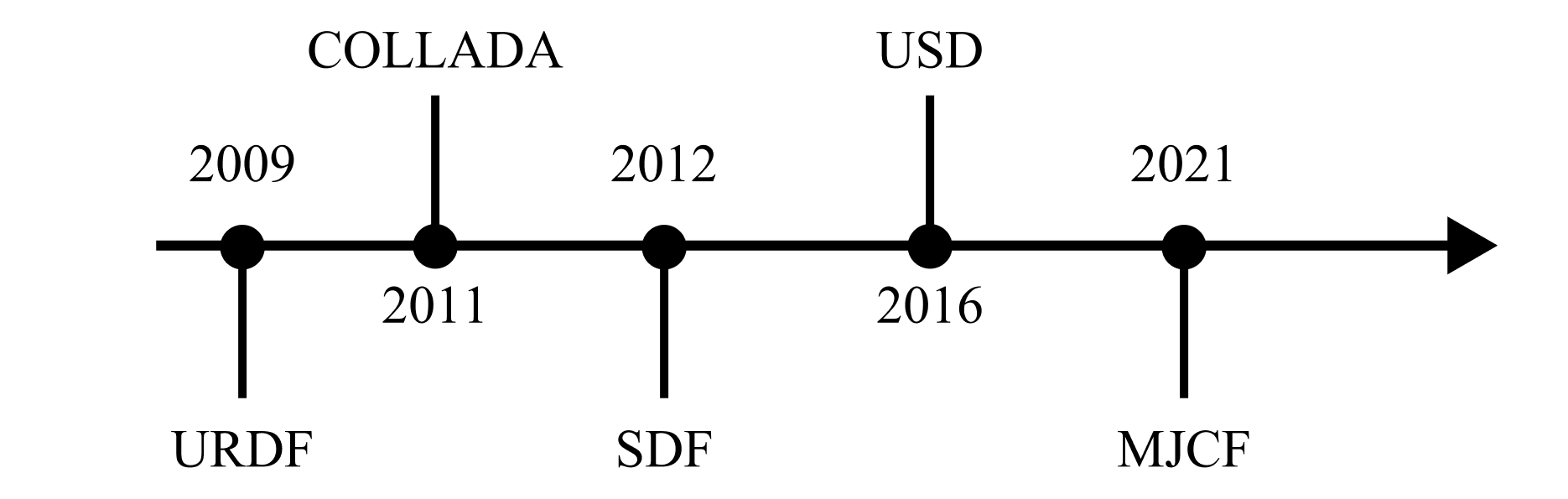}
        \caption{
        \textbf{Robot Description Format History}: 
        Timeline of the release times for each robot description format.
        }
        \label{fig:RDF-History}
    \end{figure}

    Although these robot description formats enable more comprehensive modeling information for robotic systems and have resolved some of the limitations of URDF, URDF remains the most universally adopted robot description format in academia and industry. Fig. \ref{fig:RDF-History} provides a timeline representation of the release times of these robot description formats.
    
\subsubsection{Beyond URDF}

    Daniella Tola et al. \cite{tola2023understanding-1, tola2023understanding-2} surveyed the user experience of URDF within the robotics community, including academia and industry. Their survey revealed problems associated with using URDF and inspired the research of robot description formats. Some challenges are specific to URDF, for instance, the lack of support for a closed-chain mechanism. Additionally, some challenges are common to other robot description formats, such as the complex workflow involving multiple tools, including CAD software, text editors, and simulators.

    One of the solutions is to create a new robot description format that can adequately describe robot systems and is also easy to use. A new attempt in this regard is the OpenUSD format\footnote{https://aousd.org/}, which combines the strengths of academia and industry to drive progress in this field.

    Another solution is to provide more tools to enhance the usability of robot description formats. Some tools, such as \texttt{gz-usd}\footnote{https://github.com/gazebosim/gz-mujoco/tree/main/sdformat\_mjcf} and \texttt{sdformat\_mjcf}\footnote{https://github.com/gazebosim/gz-usd}, improve the interoperability of different robot description formats. CAD tools for exporting robot designs to robot description formats are in high demand within the roboticist community because they relieve developers from the tedious workflow of creating robot description formats. Such tools include SolidWorks URDF exporter, Fusion2URDF, OnShape to URDF exporter, and the Blender extension Phobos.

    In the rest of this paper, Section \ref{sec:Methods} introduces methods for structuring the workflow from design to learning and presents an automation tool, ACDC4Robot, designed to address these challenges. Section \ref{sec:Results} demonstrates the usage of the automation tool with examples and offers a robot model library for users that can be readily utilized. We conclude in Section \ref{sec:Discuss} and discuss the limitations of our work and the future of the format for robot system development. This article's contributions include promoting a lifecycle representation from robot design to robot learning, offering the ACDC4Robot tool within Fusion 360 to streamline the workflow from robot design to robot learning, and constructing an out-of-the-box robot model library for robot design and learning.

%%%%%%%%%%%%%%%%%%%%%%%%%%%%%%%%%
\section{METHODOLOGY}
\label{sec:Methods}
%%%%%%%%%%%%%%%%%%%%%%%%%%%%%%%%%

    We analyze the workflow to describe robots from design to learning, then describe an interactive lifecycle representation. Next, we employ robot process automation to streamline the processes of robot design to robot learning. An automation tool integrated with a CAD platform can achieve this lifecycle representation interactively.

\subsection{An Interactive Lifecycle Representation}

\begin{figure}[htbp]
        \centering
        \includegraphics[width=0.7\columnwidth]{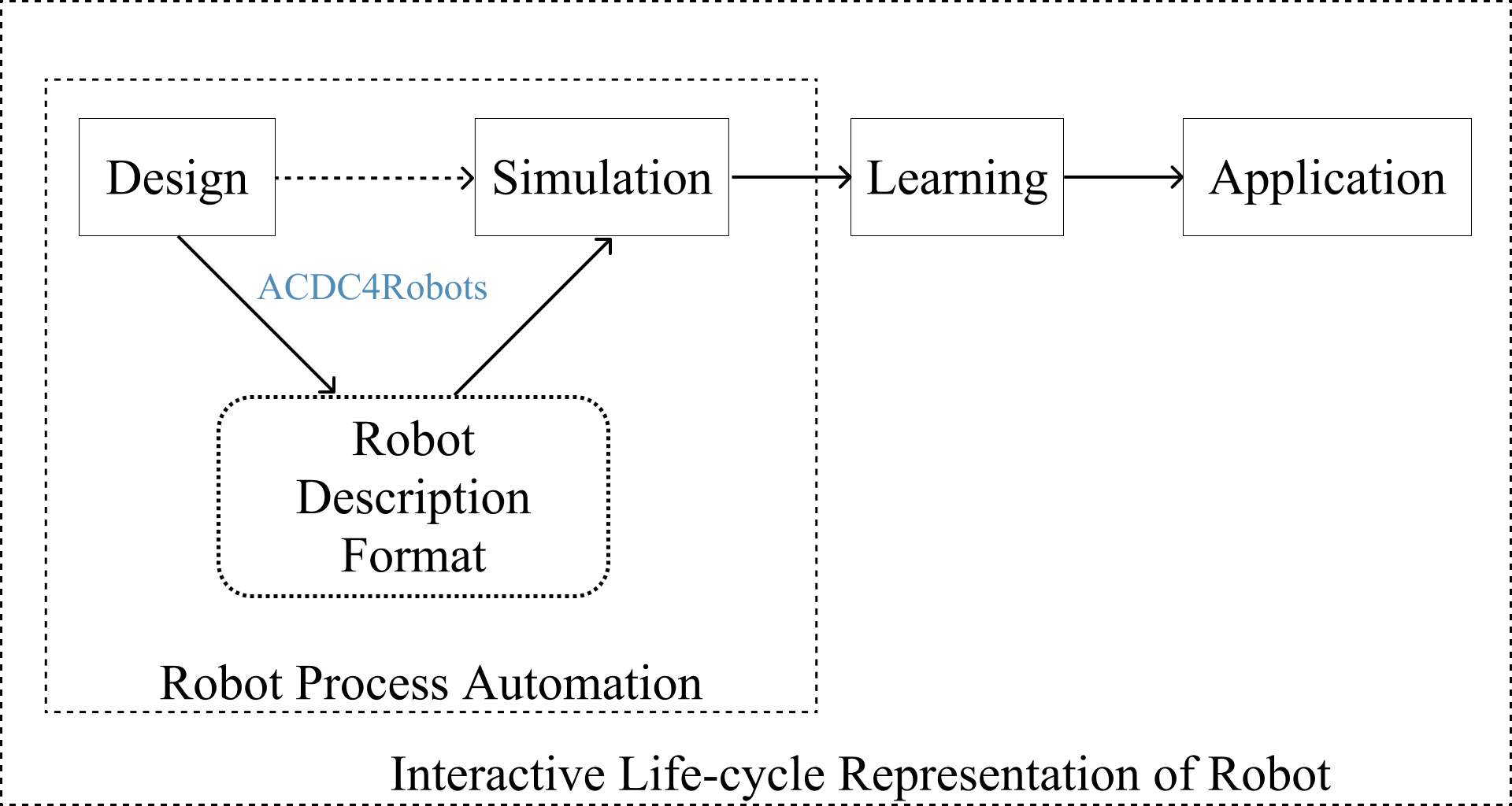}
        \caption{
        \textbf{Intercative Lifecycle Representation.}
        (i) Robot design information cannot be transferred to simulation directly due to incompatible formats;
        (ii) Design information can be converted into a robot description format through robot process automation due to the existence of a one-to-one relationship in several features between CAD and the simulator;
        (iii) An automation tool utilizes CAD software as the GUI to create and modify the robot description format, providing greater interactivity than a text editor.
        }
        \label{fig:lifecycle-representation}
    \end{figure}

    The process of robot development can be represented in various ways. Here, we separate the robot development process into four stages: design, simulation, learning, and application. In many robot learning approaches, robots are trained initially in a simulation environment and then transferred to real robots using Sim2Rreal methods. As a result, the simulation, learning, and application stages can be streamlined into a single workflow. However, the difference in file formats between the design and simulation stages poses a challenge in transferring information from robot design to simulation. To address this issue, we introduce the concept of a robot description format as a bridge to eliminate the gap between design and simulation, Fig. \ref{fig:lifecycle-representation}, allowing for the seamless connection of these stages to create a lifecycle representation of the robot development process.

    For a robot description format based on the XML format, using a text editor is a straightforward but non-intuitive method for interacting with the robot description file. Creating or modifying the robot description file by hand becomes tedious, time-consuming, and error-prone. Since the robot description format contains information directly derived from robot design, the graphical interactive interface provided by CAD software can serve as a graphical editor for the robot description format. By utilizing CAD software as the GUI, the robot description format can be interacted with in a WYSIWYG (what you see is what you get) manner. Consequently, this entire process can be regarded as an interactive lifecycle representation of the robot.

\subsection{Robotic Process Automation from Design to Simulation}

\begin{figure}[htbp]
        \centering
        \includegraphics[width=0.7\columnwidth]{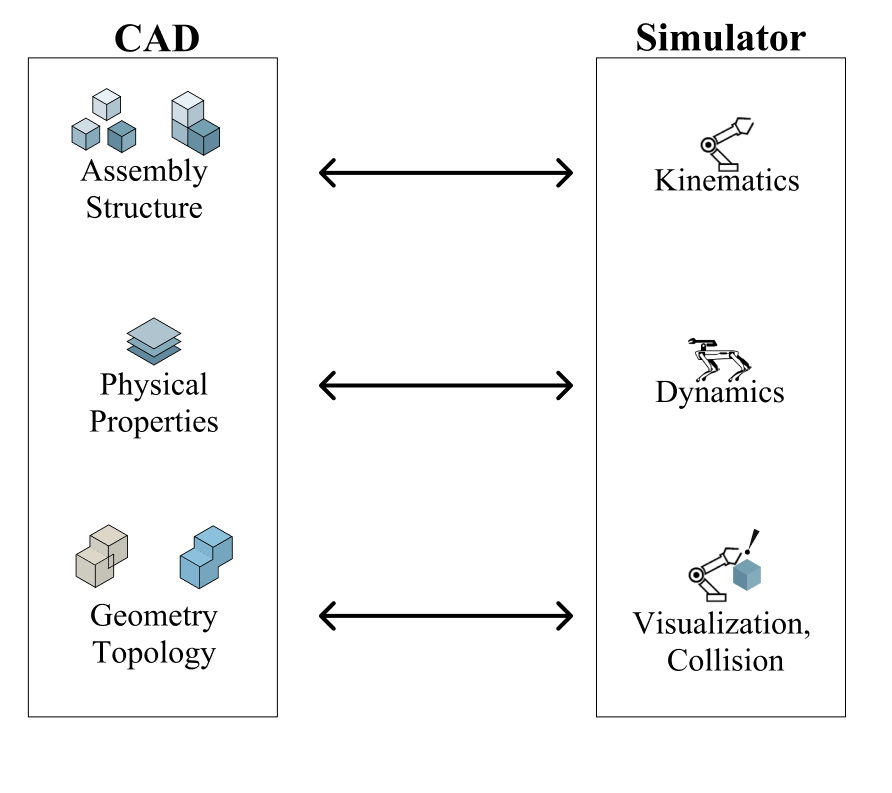}
        \caption{
        \textbf{Robot Process Automation from Design to Simulation.} 
        A one-to-one relationship exists between CAD software and robot simulators in the following features:
        (i) assembly structure and kinematics; 
        (ii) physical properties and dynamics; 
        (iii) geometry topology, visualization, and collision.
        }
        \label{fig:design-sim-mapping}
    \end{figure}

    CAD software and robot simulators are two systems with distinct functions, each emphasizing different aspects of the robot. However, some features in CAD and robot simulators represent different forms of the same information. The way components are joined to construct a robot assembly in CAD software determines the kinematics of the robot. The physical properties of robot components in CAD software can also pertain to the dynamics in the robot simulator. The geometric shape of components can be utilized for visualization and collision information in the simulator. Fig. \ref{fig:design-sim-mapping} shows that a one-to-one relationship between CAD and simulation systems enables the realization of automated conversions between these two processes, which was previously feasible.
    
\subsection{An Open-source Plug-in Using Fusion 360}

    We present an open-source plugin using Fusion 360 to achieve the interactive lifecycle process automation from robot design to robot learning. Fusion 360 is a popular CAD software developed by Autodesk within the roboticist community. It provides API access for developers, allowing it to accomplish automation tasks.

    Following J. Collins et al.'s work \cite{collins2021review}, we selected a set of popular simulators used in robotics learning, including RaiSim, Gazebo, Nvidia Isaac, MuJoCo, PyBullet, CARLA, Webots, and CoppeliaSim for comparing the compatibility of robot description formats: URDF, SDFormat, MJCF, and USD. Since we have opted to utilize Gazebo and PyBullet as our target simulation platforms, we have decided to use URDF and SDFormat according to Table \ref{tab:RDF-Comparison} as the robot description formats for transforming the design into the learning process.

    \begin{table}[]
        \caption{Comparing Simulator Support Levels with Different Robot Description Formats.}
        \label{tab:RDF-Comparison}
        \begin{center}

                \begin{tabular}{|l|l|l|}
                    \hline
                    \begin{tabular}[c]{@{}l@{}}\textbf{Robot} \\ \textbf{Description} \\ \textbf{Format}\end{tabular} & \textbf{Year} & \textbf{Supported Simulator}                                                                               \\ \hline
                    URDF                                                                & 2009 & \begin{tabular}[c]{@{}l@{}}Gazebo, Nvidia Isaac, \\ MuJoCo, PyBullet, \\ CoppeliaSim\end{tabular} \\ \hline
                    SDFormat                                                            & 2012 & Gazebo, PyBullet                                                                                  \\ \hline
                    USD                                                                 & 2016 & Nvidia Isaac                                                                                      \\ \hline
                    MJCF                                                                & 2021 & \begin{tabular}[c]{@{}l@{}}RaiSim, Nvidia Isaac, \\ MuJoCo, PyBullet\end{tabular}                 \\ \hline
                \end{tabular}
        \end{center}
    \end{table}

%%%%%%%%%%%%%%%%%%%%%%%%%%%%%%%%%
\section{RESULTS}
\label{sec:Results}
%%%%%%%%%%%%%%%%%%%%%%%%%%%%%%%%%

    In this section, we will introduce the GUI of the Fusion 360 plugin that enables the interactive lifecycle process from robot design to robot simulation, along with a guide on how to use the plugin. We will begin by using the UR5e robot manipulator model as a predefined design to verify the plugin's feasibility for exporting a serial chain robot into URDF for simulation. Next, we will employ a simple closed-loop four-bar linkage to demonstrate the workflow starting from scratch. We aim to confirm the plugin's ability to simulate closed-chain mechanisms using SDFormat in Gazebo. Additionally, we will show a robot library containing various out-of-the-box robot models to help users reduce the time spent on repetitive tasks.

    \begin{figure}[t]
        \centering
        \includegraphics[width=0.7\columnwidth]{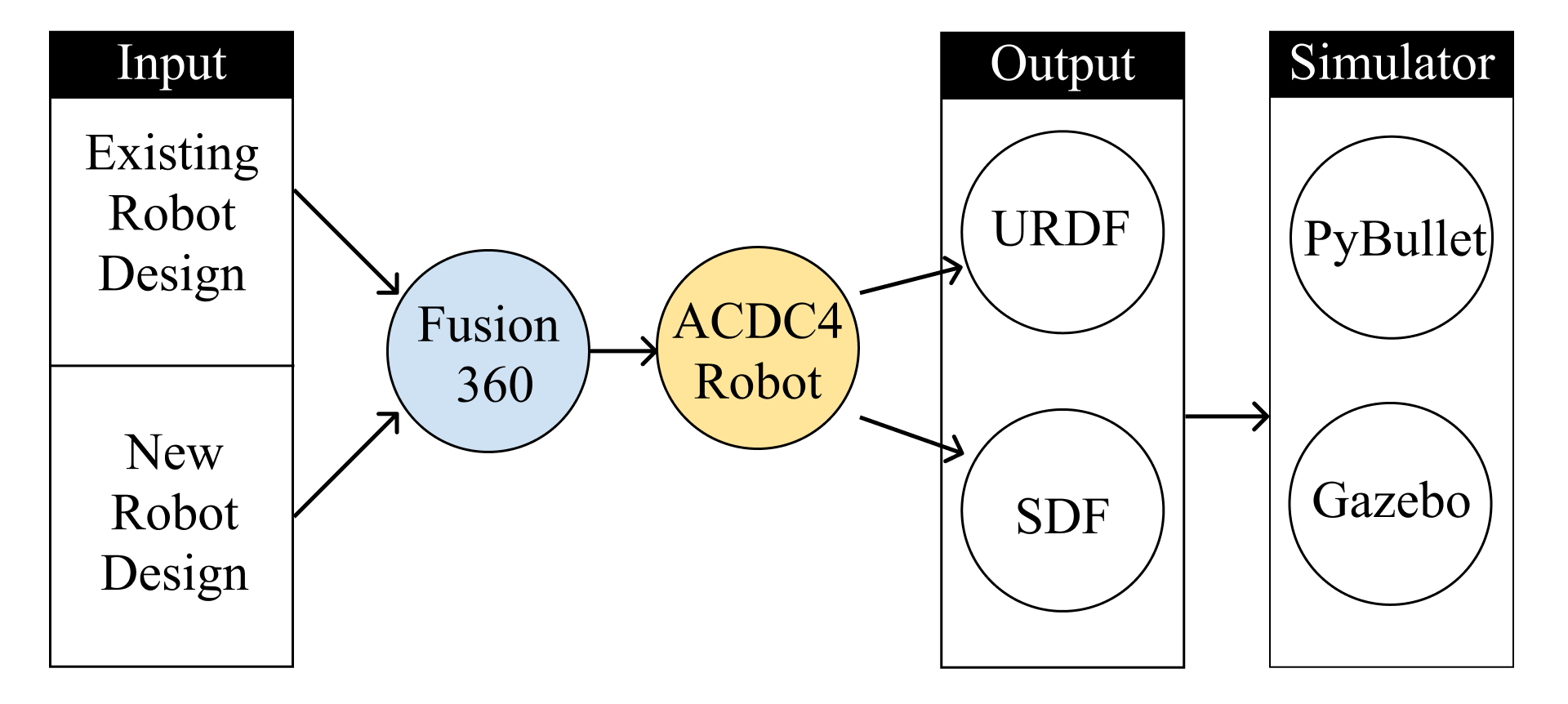}
        \caption{
        \textbf{Utilizing ACDC4Robot Plug-in in the Workflow from Design to Learning.} 
        (i) Step 1: Fusion 360 can import an existing robot model or create a new robot model from scratch;
        (ii) Step 2: ACDC4Robot plug-in converts robot design to URDF (which is not suitable for closed loop robot) or SDFormat;
        (iii) Step 3: Import the robot description format into the simulator for learning.
        }
        \label{fig:Plugin-Pipeline}
    \end{figure}

    \begin{figure}[t]
        \centering
        \includegraphics[width=0.7\columnwidth]{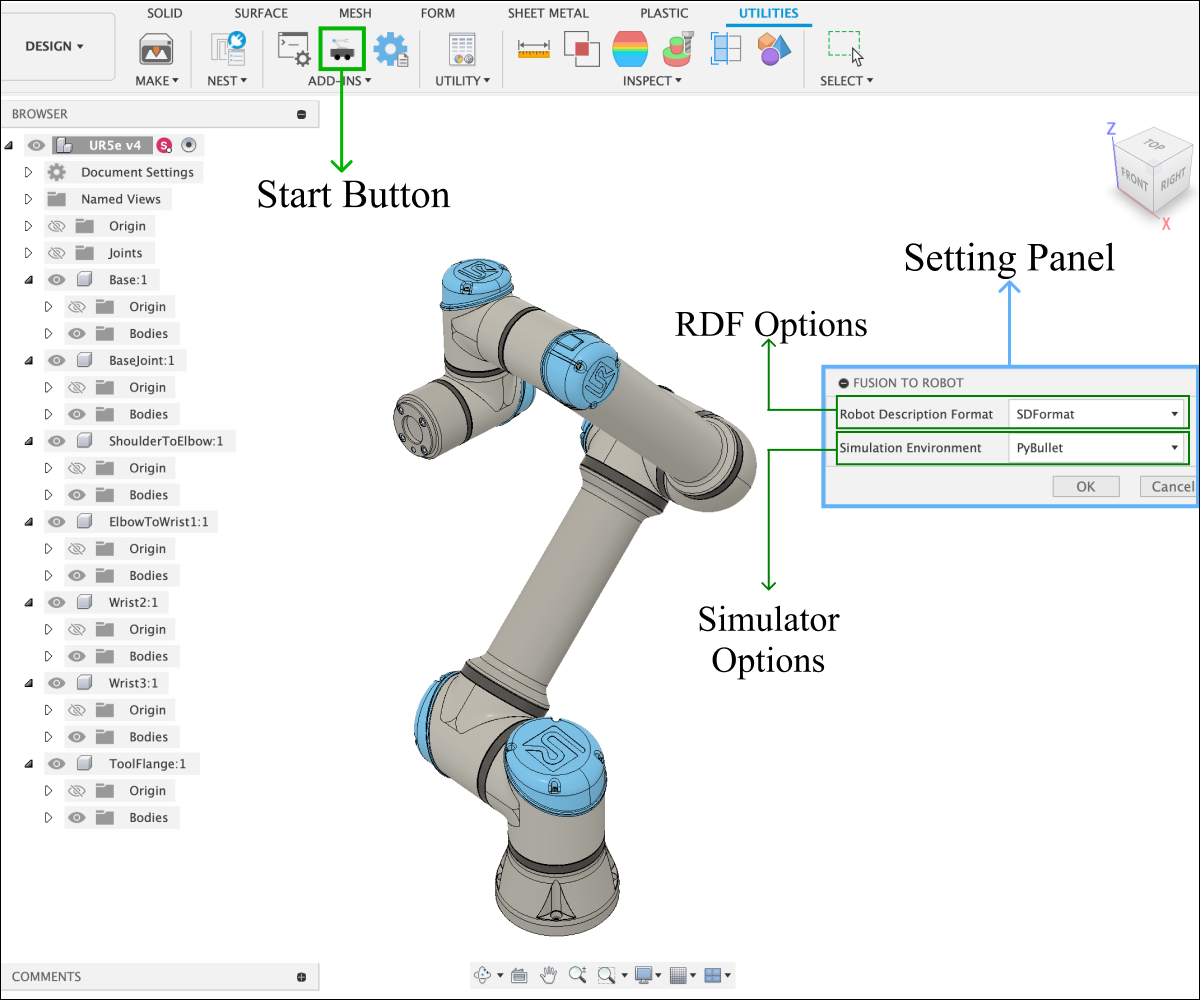}
        \caption{
        \textbf{ACDC4Robot Plugin GUI.} 
        Clicking the ACDC4Robot start button will open the settings panel, allowing the user to select the target robot description format and the target simulation environment.
        }
        \label{fig:Plugin-GUI}
    \end{figure}

\subsection{The ACDC4Robot Plug-in with Fusion 360}

    \begin{figure}[t]
        \centering
        \includegraphics[width=0.7\columnwidth]{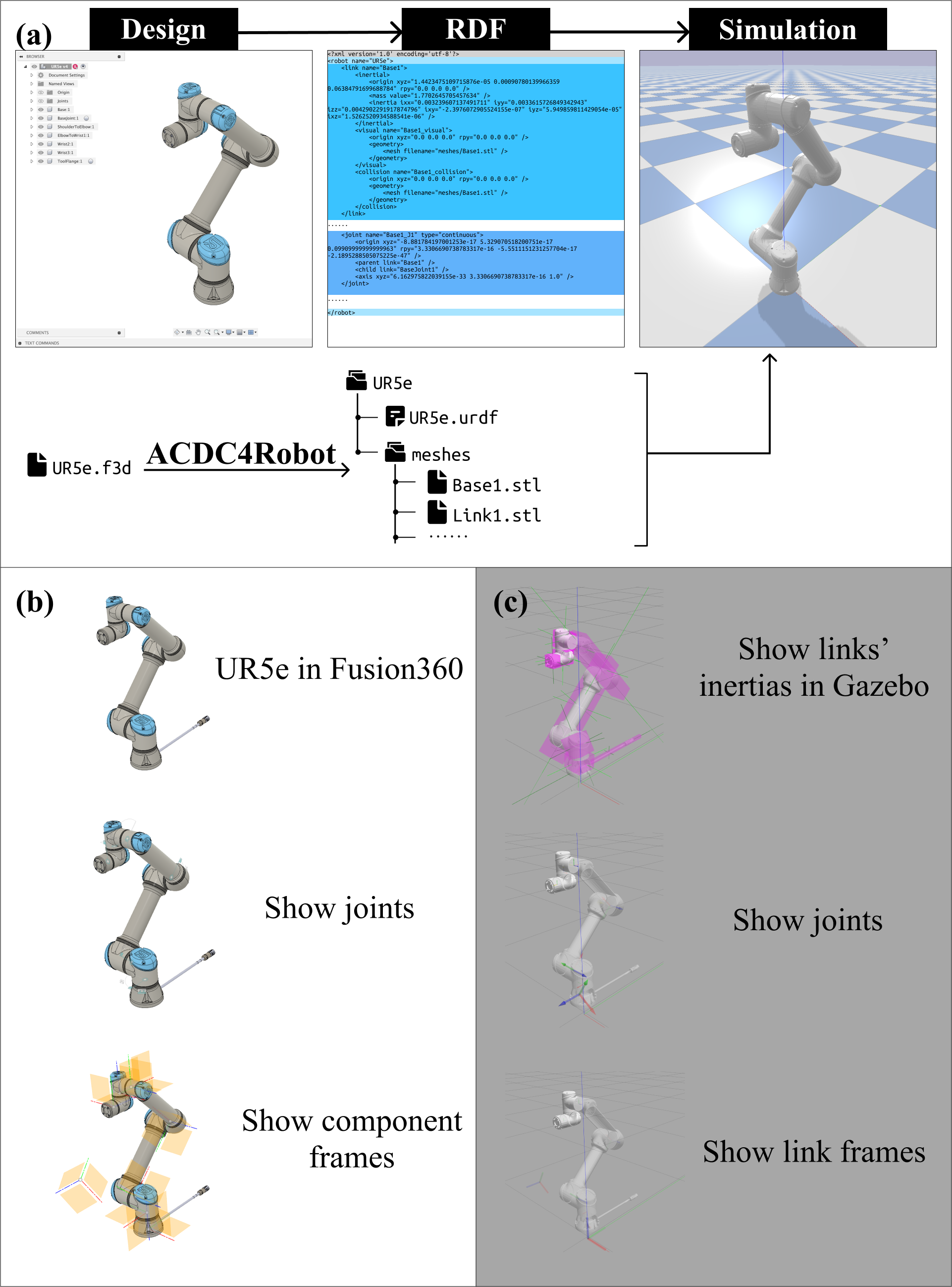}
        \caption{
        \textbf{Serial chain robot example using ACDC4Robot.} 
        (a) This figure shows the same UR5e robot manipulator model in three different representations: 
        (i) Fusion 360 design model; 
        (ii) XML-based robot description format; 
        (iii) PyBullet simulation environment for learning.
        The Fusion 360 design file \texttt{UR5e.f3d} is converted by ACDC4Robot into a \texttt{UR5e} folder that contains \texttt{UR5e.urdf} and mesh files, which are then loaded into the PyBullet simulation environment.
        (b) UR5e manipulator model and its joints, and frames in Fusion 360.
        (c) UR5e manipulator model and its joints, and link frames that come from ACDC4Robot conversion in Gazebo
        }
        \label{fig:serial-chain}
    \end{figure}

    ACDC4Robot is an open-source plugin for Fusion 360 that can automatically convert design information into a simulation data structure (robot description format) for learning. The pipeline for using the ACDC4Robot plugin from design to learning is illustrated in Fig. \ref{fig:Plugin-Pipeline}. Users can import an existing robot model into Fusion 360 or design a robot from scratch using Fusion 360. ACDC4Robot provides a straightforward GUI to simplify the conversion process, as shown in Fig. \ref{fig:Plugin-GUI}. After completing the robot morphology design, click the start button of the ACDC4Robot plugin, and a settings panel will appear on the right side of Fusion 360. This panel allows the choice between URDF or SDFormat as the format for transferring design information to the simulation and to select the target simulator for exporting in a format compatible with the chosen simulator. The exported files can then be used in the simulation for robot learning.

    Compared to the traditional method of creating and editing robot description formats using text editors, Fusion 360 as a graphical interface to modify robot design, where modifications directly reflect on the robot description file, is a more intuitive approach. Users can click a few buttons to generate robot descriptions for learning in simulation using ACDC4Robot, freeing them from the previous tedious workflow.

\subsection{Demonstrations with Serial Chains}

    It is prevalent to use existing robot models for learning purposes. Here, we use the UR5e robot manipulator model downloaded from UR's website to demonstrate the design-to-learning process with a serial chain robot, representing a typical robot type. As shown in Fig. \ref{fig:serial-chain}, the first step is to import the UR5e model into Fusion 360 and assemble it in a tree structure. Next, we use ACDC4Robot to export the design file \texttt{UR5e.f3d} into several files, including the text file \texttt{UR5e.urdf} and mesh files that are referenced in URDF as visualization and collision geometry. These robot description files can be loaded directly into PyBullet for learning.

\subsection{Demonstrations with Closed Loops}

\begin{figure}[t]
        \centering
        \includegraphics[width=0.7\columnwidth]{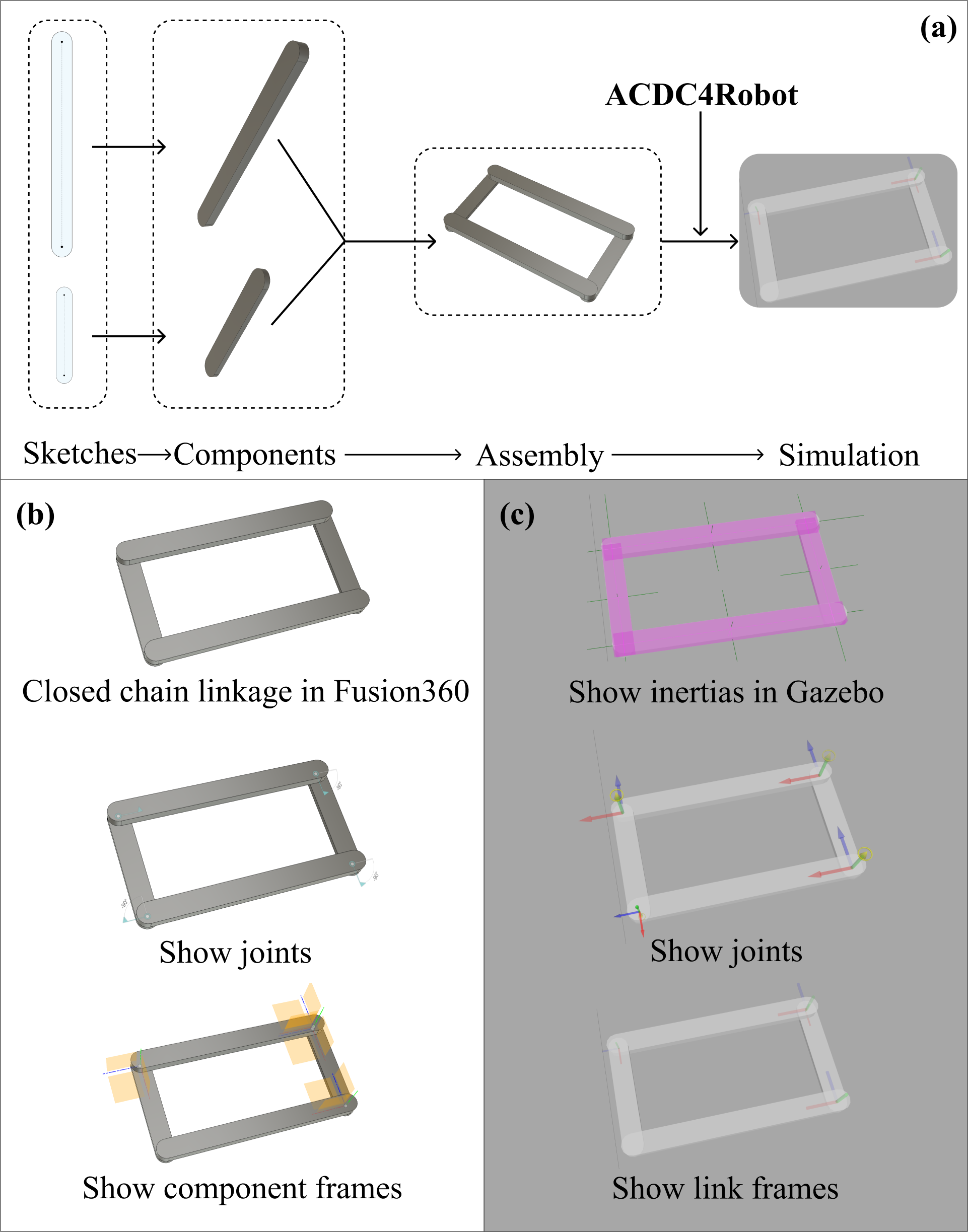}
        \caption{
        \textbf{Closed loop robot example using ACDC4Robot.} 
        (a) Create a closed-chain four-bar linkage from scratch in Fusion 360 for simulation in Gazebo;
        (b) Closed-chain four-bar linkage model and its joints, and component frames in Fusion 360;
        (c) Closed-chain four-bar linkage model and its joints, and component frames from ACDC4Robot conversion in Gazebo.
        }
        \label{fig:close-chain}
    \end{figure}

    Closed-chain mechanisms are also widely used in robots. We demonstrate creating a four-bar linkage from scratch to simulation in Fig. \ref{fig:close-chain}. First, we draw sketches for the linkage bars and then create linkage geometry components from these sketches. Next, we assemble these components by joining them together. Finally, we use ACDC4Robot to export SDFormat files (as URDF cannot describe closed-loop chain mechanisms) to Gazebo.

\subsection{Robot Library of the ACDC4Robot Plugin}

    By enabling the automatic transfer of robot design information to learning in simulation through the ACDC4Robot plugin, the robot design model becomes metadata in the interactive lifecycle representation of a robot. Consequently, a library of robot design models helps construct robot applications using the interactive lifecycle pipeline.
    
    Based on a survey of robot types modeled in URDF within the roboticist community \cite{tola2023understanding-1}, it was found that robotic arms, mobile robots, end effectors, and dual-arm robots are the most frequently used types of robots. Using this knowledge and investigating several robot datasets, including RoboSuite\footnote{https://robosuite.ai/docs/modules/robots.html}, awesome-robot-descriptions\footnote{https://github.com/robot-descriptions/awesome-robot-descriptions}, Gazebo models\footnote{http://models.gazebosim.org/}, and CoppeliaRobotics models\footnote{https://github.com/CoppeliaRobotics/models}, we have created a robot model library for Fusion 360. This library, listed in Table \ref{tab:robot-library}, has been tested with the ACDC4Robot plugin and can be used out of the box. The robot library can be downloaded from the ACDC4Robot repository.

    \begin{table}[h]
        \caption{An Robot Library for Design to Learning}
        \label{tab:robot-library}
        \begin{center}
            \begin{tabular}{|c|c|c|}
                \hline
                \textbf{Robot Name} & \textbf{Robot Type} & \textbf{Structure}\\
                \hline
                Franka Emika Panda & Robotic Arm & Serial Chain\\
                \hline
                Franka Emika Hand & End Effector & Serial Chain \\
                \hline
                Kinova Gen3 & Robotic Arm & Serial Chain \\
                \hline
                Rethink Sawyer & Robotic Arm & Serial Chain \\
                \hline
                Robotiq 2F85 Gripper & End Effector & Closed Loop \\
                \hline
                UR5e & Robotic Arm & Serial Chain \\
                \hline
                ABB YuMi & Dual Arm Robot & Serial Chain \\
                \hline
                KUKA youBot & Mobile Robot & Serial Chain \\
                \hline
            \end{tabular}
        \end{center}
    \end{table}

%%%%%%%%%%%%%%%%%%%%%%%%%%%%%%%%%
\section{DISCUSSIONS AND CONCLUSION}
\label{sec:Discuss}
%%%%%%%%%%%%%%%%%%%%%%%%%%%%%%%%%

\subsection{Towards a Lifecycle Representation}

    This article introduces an interactive lifecycle representation that spans from robot design to robot learning in simulation. We identify the gap in transferring design information to the robot simulation environment, leading us to advocate for using a robot description format to bridge the robot morphology design and robot learning in simulation. To facilitate smoother information transfer throughout the process, we have developed a robot process automation tool capable of converting design information into simulation data. This automation is made possible by the one-to-one mapping relationship between design and simulation platforms. As a result, we have created an open-source plugin called ACDC4Robot for Fusion 360. This plugin allows users to convert design information into robot description format, turning Fusion 360 into a graphical user interface (GUI) for interactive robot modeling. This interactive lifecycle process, spanning from robot design to learning, simplifies the development of robot applications.

\subsection{Limitations of the ACDC4Robot Plugin}

    Although the ACDC4Robot plugin can achieve an interactive lifecycle process from robot design to learning in simulation, it still has some limitations.

    The ACDC4Robot plugin is developed using the Fusion 360 API, making it dependent on Fusion 360. Developing a platform-independent tool capable of directly converting a design file into a robot description format for the entire lifecycle process from robot design to learning would be more versatile.

    Besides, the ACDC4Robot plugin currently only supports URDF and SDFormat. While SDFormat compensates to some extent for the limitations of URDF, such as modeling closed-chain robots, these two robot description formats can meet most of the needs of academia and industry. However, including support for exporting other robot description formats, such as MJCF, enhances the application of this tool.

    Furthermore, the number of robot models in the robot library is still relatively low compared to other robot databases. This is partly due to the challenge of obtaining publicly available robot models. Additionally, assembling these acquired robot models in Fusion 360 and testing them with the ACDC4Robot plugin is quite time-consuming. We plan to incrementally expand the robot library during the future development process.

\subsection{Towards a Unified Robot Lifecycle Format}

    The modularity of the robot development process has led to separate formats for storing information, creating barriers to data exchange between different stages of development. While conversion tools have partially addressed this issue, they must improve efficiency. An ultimate solution would involve adopting a universal format describing all the information across the robot development lifecycle. The adoption of such a universal format has the potential to enhance the efficiency of robot development significantly. The Universal Scene Description (USD) format is moving in this direction.

%%%%%%%%%%%%%%%%%%%%%%%%%%%%%%%%%%%%%%%%%%%%
\section*{Acknowledgements}
%%%%%%%%%%%%%%%%%%%%%%%%%%%%%%%%%%%%%%%%%%%%

    This work was partly supported by the Ministry of Education of China-Autodesk Joint Project on Engineering Education, the National Natural Science Foundation of China [62206119], and the Science, Technology, and Innovation Commission of Shenzhen Municipality [JCYJ20220818100417038, ZDSYS20220527171403009, and SGDX20220530110804030].

%%%%%%%%%%%%%%%%%%%%%%%%%%%%%%%%%%%%%%%%%
\bibliographystyle{unsrt}
\bibliography{References}  %%% Remove comment to use the external .bib file (using bibtex).
%%%%%%%%%%%%%%%%%%%%%%%%%%%%%%%%%
\end{document}